\documentclass[10pt, a4paper]{article}

\usepackage[final]{lrec2026} 

\usepackage{svg}
\usepackage[svgnames]{xcolor}
\usepackage{amsmath}
\usepackage{float}
\usepackage{booktabs}
\usepackage[table]{xcolor}
\usepackage{ulem}
\usepackage{enumitem}
\usepackage{tcolorbox}
\usepackage{fancyhdr}
\usepackage{tikz-dependency}
\tcbuselibrary{listings}

\usepackage{hyperref}

\title{Profiling Hallucinations in Frontier LLMs for \\Entity Linking to Medical Ontologies}

\name{Logan Born$^1$, Nishant Kambhatla$^1$, Uliyana Kubasova$^1$, Maryam Siahbani$^{1,2}$, \\ {\bf \large Andrei Vacariu$^1$, Tim O'Connell$^1$ and Anoop Sarkar$^1$}}
\address{
$^1$Emtelligent \qquad $^2$University of the Fraser Valley \\
\{lborn,nishant,uliyana,maryam,andrei,tim,anoop\}@emtelligent.com
}

\abstract{
The integration of Large Language Models (LLMs) into healthcare promises to revolutionize clinical documentation and interoperability, yet reliability remains a concern. This study presents a comprehensive analysis of hallucinations by frontier LLMs tasked with mapping clinical text to SNOMED CT. Through rigorous experimentation, we identify a critical reliability gap: LLMs hallucinate medical codes at a rate that currently renders them unsuitable for autonomous clinical coding applications. Paradoxically, constraining models to use ground-truth mention spans exacerbates, rather than mitigates, these hallucinations. We further contribute a taxonomy of hallucination types -- including deprecated codes and cross-ontology errors -- and demonstrate that general-purpose LLMs significantly underperform compared to specialized zero-shot entity linking approaches. These findings underscore the need for robust verification mechanisms before clinical deployment.
 \\ \newline \Keywords{clinical nlp, snomed-ct, ontologies, entity linking, LLMs, hallucination} }

\begin{document}

\maketitleabstract

\section{Introduction}

The transition from specialized discriminative models to general-purpose LLMs has prompted an ongoing re-evaluation of clinical NLP pipelines~\cite{Shool2025,Maity2025-yt}. While frontier LLM providers such as Anthropic\footnote{\url{https://www.anthropic.com/news/healthcare-life-sciences}} and OpenAI\footnote{\url{https://openai.com/index/openai-for-healthcare/}} advocate for LLMs as end-to-end agents in healthcare settings, the overall utility of these models depends heavily on their ability to ground their output into established knowledge bases; in clinical settings, this grounding may be achieved through entity linking to an ontology such as SNOMED CT.

\begin{figure}[t]
    \centering
    \includegraphics[width=0.9\columnwidth]{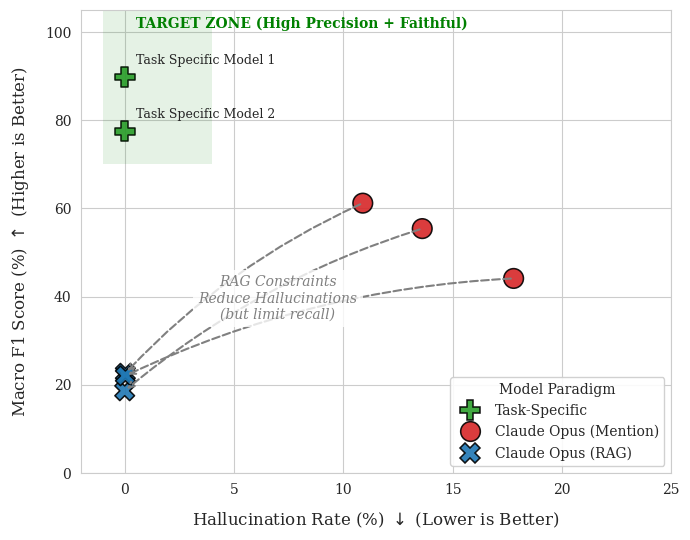}
    \caption{Comparison of generalist and specialist models on a clinical entity linking task. 
    Task-specific architectures like KIRI \cite{snomedct-challenge} achieve high extraction performance without hallucinating. In contrast, Generalist LLMs exhibit a strict trade-off: they either suffer from high hallucination rates in end-to-end settings or suffer from low recall in RAG settings, failing to simultaneously optimize for both faithfulness and accuracy.}
    \label{fig:heatmap}
\end{figure}
            
However, treating medical entity linking as a generative task introduces a novel failure mode absent in discriminative baselines: structural hallucination~\cite{10.1145/3703155}. Unlike traditional classification models that select from a fixed candidate set, generative models construct identifiers token-by-token. This creates the possibility of generating plausible-looking but non-existent identifiers. Evaluating these hallucinations in open-domain text is notoriously difficult due to the ambiguity between factual errors and legitimate inference (ibid.).

Medical entity linking against an ontology such as SNOMED CT offers a uniquely tractable testbed for quantifying LLM faithfulness by facilitating hallucination detection. SNOMED CT Identifiers (SCTIDs) are not arbitrary strings: they belong to a strictly controlled vocabulary with internal structure and built-in validation features.\footnotemark These properties allow us to disentangle \textit{alignment errors}, where the model retrieves an incorrect but valid concept, from \textit{hallucinations}, where a model fabricates a code that is structurally invalid or non-existent.\footnotetext{{\url{https://docs.snomed.org/snomed-ct-specifications/snomed-ct-release-file-specification/snomed-ct-identifiers/6.4-check-digit}}}

We investigate the capacity of frontier LLMs to perform mention-based linking against SNOMED CT, comparing them against state-of-the-art discriminative architectures. Our experiments reveal significant shortcomings in LLMs' ontological reasoning abilities (Figure~\ref{fig:heatmap}):
\\\textbf{High Hallucination Rates:} In standard prompting setups, frontier LLMs fail to respect the closed vocabulary requirement, with 20--55\% of generated codes being invalid.
\\\textbf{Precision-Recall Tradeoff:} Retrieval-Augmented Generation (RAG) mitigates hallucination, but also severely bottlenecks recall. Generalist models struggle to synthesize retrieved information and fail to match task-specific model performance.
\\\textbf{Complex Failure Modes:} We profile errors from generative models, which include the mixing of disparate ontology extensions (e.g., UK vs.\ US editions) and the resurrection of deprecated codes.

Ultimately, our results suggest that medical coding remains a challenging domain where specialized, discriminative architectures significantly outperform generalist generative approaches in both safety and accuracy.

\section{Methodology}
We propose to prompt three top-of-the-line large language models (Claude Sonnet 4.5, Claude Opus 4.6, and GPT-5.2) to perform entity linking tasks on a corpus of discharge summaries. We aim first to quantify how frequently these models hallucinate in their output: of the concept IDs which they produce, what proportion really exist in the US release of the SNOMED CT ontology? Secondly, we aim to describe qualitatively \textit{how} these models hallucinate: what proportion of their outputs are outdated, invalid, from nonstandard ontology extensions, or are pure fabrications.

We aim to show that, despite advances in grounding LLM outputs through techniques like retrieval-augmented generation, even frontier models remain prone to fabricating outputs. While some of these hallucinations are immediately obvious, and would likely be caught and pruned through a simple post-hoc filtering step, others are more subtle and may not be immediately evident as errors.

\section{Experimental Setup}
\label{sec:experiments}

\subsection{Data}\label{sec:data}

We use a revised version of the SNOMED CT Entity Linking Challenge dataset~\cite{PhysioNet-snomed-ct-entity-challenge-1.1.0}, which comprises 272 discharge summaries from MIMIC-IV~\cite{PhysioNet-mimic-iv-note-2.2}. 

We manually update the annotations in the original dataset using the 2025-03-01 US Edition of SNOMED CT, which combines the International Release with the US Extension.
Additionally, we re-annotate these documents using a commercial medical NLP engine (Appendix~\ref{sec:datacleanup}).
The original dataset labels some phrases inconsistently, such that (for example) \texttt{BLOOD Glucose} is occasionally labelled as \texttt{Glucose measurement, blood} and identical strings in other settings are unlabelled (see~\citealt{snomedct-challenge} for additional discussion). The original dataset also intentionally excludes some SNOMED CT sub-hierarchies. Our reannotation efforts are intended to make the dataset more consistent and to provide coverage for more parts of the SNOMED CT ontology.

We prompt Claude Haiku\footnote{We use a small model for this task as we believe this type of pass/fail judgment does not require the full reasoning power of a larger model.} 4.5 to act as an independent judge, filtering the NLP engine outputs by flagging any that are incorrect (either inherently, or due to some feature of the surrounding context), irrelevant, or too broad or narrow. For example, in ``\texttt{pt was initially oriented x1}'', ``\texttt{oriented}'' is assigned the concept \texttt{Oriented to person, time and place (finding)}, when the context implies only a single degree of orientation; the LLM  therefore filters this annotation from the final test set. We take the union of any surviving annotations with the annotations from the original dataset; new annotations take priority in case of overlap.

The resulting dataset comprises the same 272 documents as the original (68 test, 204 train) and a total of 
78961 annotated links to SNOMED concepts (24087 test annotations, 54874 train annotations). It covers 6271 distinct SNOMED concepts (5151 in training, 3845 in test; 1120 concepts are zero-shot, occurring only in the training set), with a minimum frequency of 1, maximum of 786, and median of 2.\footnote{Upon request, we will share a patch allowing parties with access to the original SNOMED CT Entity Linking Challenge dataset to update to our revised version.}

This reannotation effort clearly biases any entity linking evaluation in favour of our own model. However, the intended focus of this work is on analyzing and taxonomizing hallucinations; we are not primarily interested in entity linking accuracy \textit{per se}, and we therefore consider this an acceptable bias. We include our model in the tables of results to show the expected upper bound of performance achievable when there is perfect alignment between the test set and a model's level of detail and phrase segmentation choices, but grey it out to highlight that it is not a fair comparand.

\subsection{Baselines}

To rigorously evaluate the capabilities of frontier LLMs in clinical concept normalization and to isolate the sources of hallucination, we devised four distinct experimental configurations. These settings range from isolating the linking capability to evaluating full pipeline performance and testing retrieval-augmented strategies.

\paragraph{Oracle Span} In this setting, we decouple entity recognition from entity linking. The LLM is provided with the clinical note text alongside the ground-truth character offsets for target entities. The model is tasked with predicting the correct SNOMED CT concept ID for each provided span. To manage context window limitations and optimize throughput, entities are processed in batches (50 per inference call). This setting serves as an upper bound for the model’s internal parametric knowledge, isolating its ability to map clinical language to ontology codes without the confounding effects of extraction errors.

\paragraph{End-to-End (E2E)} This is the most stringent evaluation, simulating a raw deployment scenario. The LLM is provided only with the raw clinical text and must perform joint Named Entity Recognition and Normalization (NER+N). The model is tasked with extracting entity mentions, determining their exact character start and end offsets, and predicting the associated SNOMED CT code. While this tests the viability of a monolithic LLM pipeline, it penalizes models for tokenization artifacts; LLMs relying on sub-word tokenizers often struggle with precise character counting, resulting in scoring failures despite semantically correct extraction.

\paragraph{Mention-based Extraction (Mention)} To mitigate errors from imprecise character counting while still evaluating extraction capabilities, we attempt a Mention-based approach. The LLM is instructed to extract entities by providing the surface text and occurrence count, for example, \texttt{("hypertension", 3)} denotes the third instance of the string "\texttt{hypertension}". We resolve these predictions to specific spans via post-hoc programmatic resolution using case-insensitive string matching. This setting allows us to assess the model's ability to identify and normalize clinical concepts based on semantic context, independent of its ability to perform arithmetic character indexing.

\paragraph{Retrieval Augmented Generation (RAG)} We additionally implement a RAG pipeline that moves the burden of knowledge retrieval from the LLM’s parametric memory to an external index. This follows a three-step pipeline:

\begin{enumerate}[itemsep=2pt]
    \item \textbf{Extraction} The LLM extracts mentions (surface form and occurrence index) from the raw text, but does not predict any concept IDs.
    \item \textbf{Retrieval} Extracted mentions are converted into embedding vectors and queried against a database of pre-computed embeddings of SNOMED CT Fully Specified Names (FSNs). We retrieve the top 5 candidate concepts based on cosine similarity. OpenAI's \texttt{text-embedding-3-small} is used to build $512$ dimensional embeddings \footnote{While the hyperparameter space for RAG (e.g., retrieval depth $k$, re-ranking strategies) is extensive, we fix $k=5$ for all experiments. Given the significant length of clinical notes, a smaller $k$ minimizes context dilution and ensures the LLM remains focused on the most relevant candidates. Furthermore, our primary objective is to evaluate the architecture's impact on hallucination mitigation rather than to optimize retrieval performance.}. 
    \item \textbf{Selection} The LLM is presented with the extracted mention and the retrieved candidate list, and must select the most appropriate concept ID. By constraining the LLM’s output space to retrieved candidates, this setting aims to minimize hallucinations of non-existent or irrelevant medical codes.
\end{enumerate}

We evaluate three state-of-the-art generalist models: Claude Opus 4.6, Claude Sonnet 4.5, and GPT-5.2. To ensure a robust comparative analysis, we employ a full factorial experimental design, evaluating every model across all four settings (Oracle Span, E2E, Mention, and RAG). This comprehensive evaluation allows us to directly compare the impact of architectural differences against methodological constraints, disentangling the models' intrinsic reasoning capabilities from the difficulty of the extraction and retrieval tasks.

To put the results in perspective, we compare against a closed-source commercial entity linking system called \textit{entity-paradigm},
against AWS Medical Comprehend\footnote{\url{https://aws.amazon.com/comprehend/medical/}}, and against KIRI, an open-source dictionary-based system which was the winning submission to the original SNOMED CT entity linking challenge~\cite{snomedct-challenge}. These models do not generate codes from outside the target ontology, and therefore serve as a hallucination-free baseline. We retrain KIRI on the updated training annotations; the other comparands are zero-shot and see only the test set.

We report micro- and macro-averaged hallucination rates for all settings, which we define as the proportion of concept IDs in the output which do not exist in the 2025-03-01 US release of SNOMED CT, which was the current release at the time of these experiments.

We report macro-averaged character IoU using the official scoring script \cite{snomedct-challenge} in keeping with prior work on the SNOMED CT Entity Linking Challenge dataset. We additionally report precision, recall, and macro-F1 for settings where the model must extract mentions prior to linking (discriminative model, E2E LLM, and the LLM Mention setting). For the LLM Oracle Span setting, where the model is provided the spans, we instead report the accuracy of the assigned concept links.

\section{Results}

\begin{table*}[htbp]
    \centering
    \scalebox{0.85}{
    \begin{tabular}{l @{\hskip 1cm} rrrrrrr}
          \toprule
        
        \multicolumn{7}{c}{
        \textbf{Task-Specific Models}
        }\\\midrule

          \textbf{Model} 
        & \textbf{IoU $\uparrow$} 
        & \textbf{P $\uparrow$} 
        & \textbf{R $\uparrow$} 
        & \textbf{F1 $\uparrow$} 
        & \textbf{H $\downarrow$} 
        & \textbf{Macro H $\downarrow$} 
        \\\midrule
        
        emtelligent entity-paradigm (Ours) 
        & \color{lightgray}\dashuline{78.06\%} 
        & \color{lightgray}\dashuline{91.95\%} 
        & \color{lightgray}\dashuline{87.84\%} 
        & \color{lightgray}\dashuline{89.85\%} 
        & \dashuline{0.00\%} 
        & \dashuline{0.00\%} 
        \\
        KIRI \cite{snomedct-challenge} 
        & 50.49\% 
        & 77.33\% 
        & 77.91\% 
        & 77.61\% 
        & \dashuline{0.00\%} 
        & \dashuline{0.00\%} 
        \\\midrule

        Amazon Comprehend Medical$^\dagger$ 
        & 22.06\%
        & 40.22\%
        & 49.91\%
        & 22.37\%
        & \dashuline{0.00\%}
        & \dashuline{0.00\%}
        \\
        \toprule

        \multicolumn{7}{c}{
        
        \textbf{General-Purpose Frontier LLMs}}\\\midrule

          \textbf{Model} 
        & \textbf{IoU $\uparrow$} 
        & \textbf{P $\uparrow$} 
        & \textbf{R $\uparrow$} 
        & \textbf{F1 $\uparrow$} 
        & \textbf{H $\downarrow$} 
        & \textbf{Macro H $\downarrow$} 
        \\\midrule
        
        GPT-5.2 (E2E) 
        & 0.15\% 
        & 58.79\% 
        & 9.46\% 
        & 16.30\% 
        & 12.21\% 
        & 19.47\%
        \\
        Claude Opus 4.6 (E2E) 
        & 2.23\%
        & 2.34\%
        & 2.95\%
        & 2.61\%
        & 13.61\%
        & 42.21\%
        \\
        Claude Sonnet 4.5 (E2E) 
        & 1.60\%
        & 1.56\%
        & 2.15\%
        & 1.81\%
        & 18.84\%
        & 55.04\%
        \\\midrule
        
        GPT-5.2 (Mention) 
        & 5.39\% 
        & \textbf{72.25\%} 
        & 44.97\% 
        & 55.44\% 
        & 13.61\% 
        & 28.31\%
        \\
        
        Claude Opus 4.6 (Mention) 
        & 8.98\% 
        & 68.48\% 
        & \textbf{55.33\%} 
        & \textbf{61.21\%} 
        & 10.89\% 
        & 24.94\%
        \\
        
        Claude Sonnet 4.5 (Mention) 
        & 4.78\% 
        & 72.20\% 
        & 31.79\% 
        & 44.14\% 
        & 17.79\% 
        & 34.53\%
        \\
        \midrule

        GPT 5.2 (RAG)
        & 11.66\%
        & 15.59\%
        & 23.27\%
        & 18.67\%
        & \textbf{0.00\%}
        & \textbf{0.00\%}
        \\
        Claude Opus 4.6 (RAG)
        & \textbf{14.56\%}
        & 20.37\%
        & 25.47\%
        & 22.64\%
        & 0.003\%
        & 0.02\%
        \\
        Claude Sonnet 4.5 (RAG)
        & 14.08\%
        & 19.62\%
        & 25.19\%
        & 22.06\%
        & 0.003\%
        & 0.02\%
        \\
        \bottomrule
    \end{tabular}
    }
    \caption{Macro-averaged Intersection over Union (IoU), precision (P), recall (R), F1 (F1), hallucination rate (H), and macro-averaged hallucination rate (Macro H) on SNOMED CT Entity Linking Challenge; KIRI is the 2025 winning submission \cite{snomedct-challenge}. \hfill$^\dagger$AWS Comprehend score uses top-5 outputs. Others use top-1 output.}
    \label{tab:snomedct-challenge-mention}
\end{table*}

\begin{table*}[htbp]
    \centering
    \scalebox{0.85}{
    \begin{tabular}{l @{\hskip 1cm} rrrrrr}
        \toprule
        \textbf{Model} 
        & \textbf{IoU $\uparrow$} 
        & \textbf{H $\downarrow$} 
        & \textbf{Macro H $\downarrow$} 
        & \textbf{Acc (Real) $\uparrow$} 
        & \textbf{Acc (Overall) $\uparrow$}
        \\\midrule
        emtelligent entity-paradigm (Ours)
        & \color{lightgray}84.45\%
        & \color{lightgray}0\%
        & \color{lightgray}0\%
        & \color{lightgray}87.84\% 
        & \color{lightgray}87.84\%
        \\\midrule
        
        GPT-5.2 (Oracle Span) 
        & 6.69\%
        & 14.28\%
        & 38.19\%
        & 34.79\%
        & 29.83\% \\

        Claude Opus 4.6 (Oracle Span)
        & 12.36\%
        & 13.57\%
        & 34.47\%
        & 52.43\%
        & 45.31\% \\
        
        Claude Sonnet 4.5 (Oracle Span) 
        & 8.02\%
        & 19.34\% 
        & 52.50\% 
        & 48.11\% 
        & 38.81\% 
        \\
        \bottomrule
    \end{tabular}
    }
    \caption{Macro-averaged Intersection over Union (IoU), hallucination rate (H), macro-averaged hallucination rate (Macro H), accuracy limited to SCTIDs which actually exist (Acc (Real Only)) and overall accuracy (Acc (Overall)) on SNOMED CT Entity Linking Challenge when models are provided oracle access to the ground truth span boundaries.}
    \label{tab:snomedct-challenge-span}
\end{table*}

\subsection{Hallucination Rate}
In settings without RAG, the micro-averaged hallucination rate, i.e. the overall proportion of concept ID tokens which do not represent a real SNOMED concept, ranges from a minimum of 10.89\% in the Claude Opus (Mention) setting to a maximum of 19.34\% from Claude Sonnet (Oracle Span).

The macro-averaged hallucination rate, i.e. the proportion of \textit{distinct} concept ids which are not real SCTIDs, is much higher, between 19.47\% in the GPT-5.2 (End-to-end) setting and 52.50\% in the Claude Sonnet (Oracle Span) setting. 

As may be expected, RAG significantly reduces the hallucination rate for all models, though notably it does not eliminate hallucinations entirely: both Claude Opus and Sonnet hallucinate a single SCTID even in the RAG setting.

\subsection{Character IoU and F1}
As seen in Tables~\ref{tab:snomedct-challenge-mention} and \ref{tab:snomedct-challenge-span}, generalist LLMs fail to achieve competitive macro-averaged character IoU scores when compared to task-specific models.

The end-to-end approach using GPT-5.2 performs particularly poorly, achieving just 0.015 IoU. This setting requires the model to not only produce accurate concept labels, but also to accurately count the character spans associated with these labels, a challenging task for the LLMs.

Scores in the Mention setting, where the model is required to perform entity extraction and linking, are better, though still do not exceed 8.98\%.

The Oracle Span setting, which tests models only on their ability to assign correct concept labels, produces the highest IoU of any LLM-based approach, though even this score of 12.36\% falls well short of the scores achieved by task-specific methods.

In both the E2E and Mention settings, precision generally outpaces recall, implying that all models regardless of provider fail to extract a majority of possible mentions. Recall is particularly poor in the E2E setting, where more than 90\% of mentions go undiscovered, but even the best-case (Claude Opus, Mention) misses nearly half of all mentions.

Only for the metric of precision do LLM-based techniques begin to approach dictionary-based methods, where the best precision (72.25\%, achieved by GPT-5.2, Mention) is just 5 points short of the prior state-of-the-art on this dataset.

Unlike the other models in this comparison, note that AWS Medical Comprehend outputs a top-$k$ list of candidate concepts. Table~\ref{tab:snomedct-challenge-mention} reports top-5 scores for this model.
We note that the top-1 prediction is often semantically valid but differs in scope from the ground truth, e.g. \texttt{Adverse reaction} when the desired target is \texttt{Adverse reaction to drug}.
A top-1 evaluation penalizes these cases, leading to a very low F1 score of just 0.25\%.
Performing a top-5 evaluation reduces the penalty for near-misses of this sort, as the broader or narrower target concept is often present in the list of outputs, albeit with a lower score.

\section{Analysis}

\subsection{Hallucination Rate}
\begin{figure}[b]
    \centering
    \includegraphics[width=\linewidth]{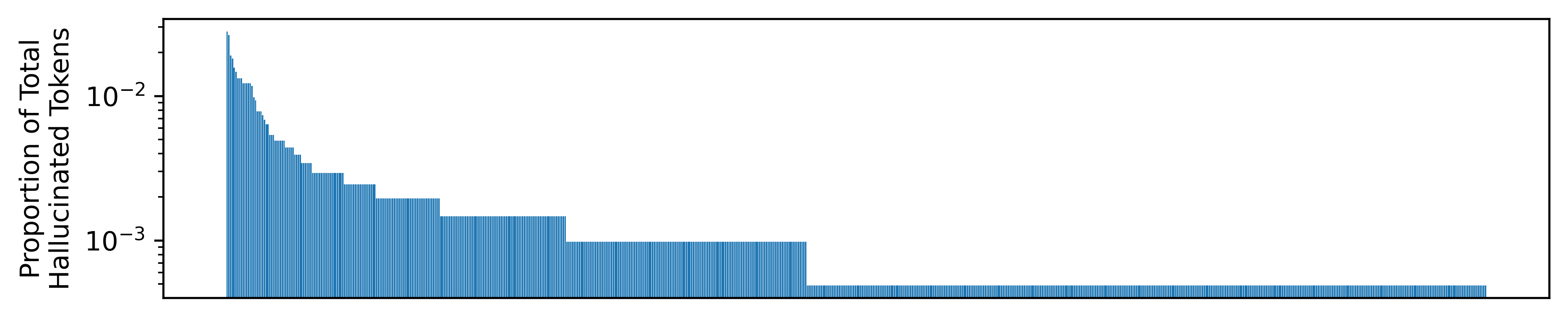}
    \includegraphics[width=\linewidth]{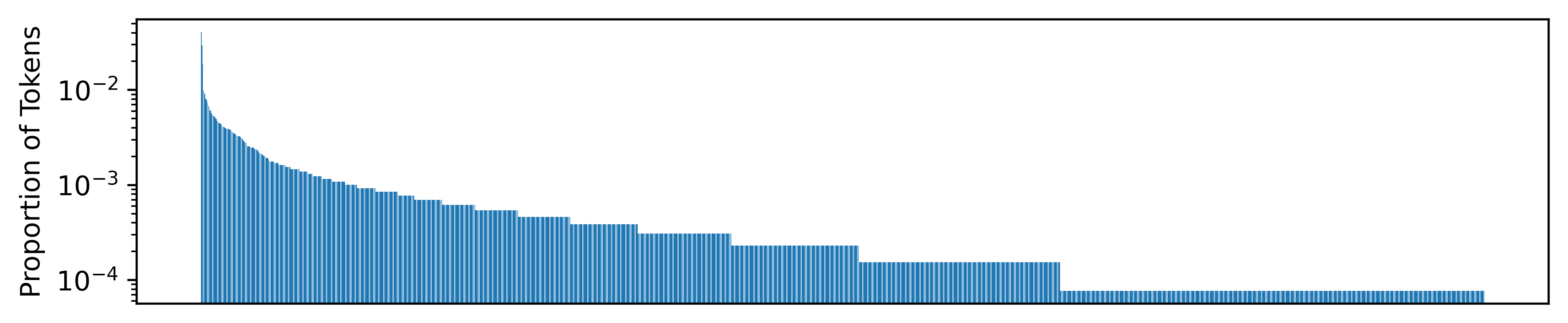}
    \caption{Rate of repetition of hallucinated codes (top) and real codes (bottom) in the GPT-5.2 (Mention) setting; each bar represents one concept id. Other settings exhibit a similarly long tail of rare and one-off codes.}
    \label{fig:repetition_rate}
\end{figure}

The asymmetry between micro- and macro-averaged hallucination rates implies that models do not consistently hallucinate the same codes many times. Rather, there are a small number of codes which are hallucinated frequently, followed by a wide range of codes attested a few times apiece. This is visualized in Figure~\ref{fig:repetition_rate} (top), which shows the very long tail of rare hallucinations produced in the GPT-5.2 (Mention) setting.
Note that frequency alone is not a sufficient condition for identifying hallucinations: real concept ids follow a similarly long-tailed distribution (Figure~\ref{fig:repetition_rate}, bottom).

\subsection{Impact of Frequency}
There is, at best, a very weak correlation between concept frequency and a model's tendency to hallucinate. We estimate the overall frequencies of SNOMED CT concepts by annotating 100k discharge summaries using a commercial medical NLP engine and counting the number of occurrences of every SCTID in the resulting output. We measure Pearson's $\rho$ between a concept's frequency in this dataset and the proportion of outputs that were hallucinated in the Oracle Span setting when the target label equalled the concept in question. We find $\rho = -0.0386$, $p = 0.0007$, i.e. there is a significant but extremely weak tendency for rarer concepts to induce more hallucinations.

There is no significant correlation ($\rho = -0.00642$, $p = 0.2308$) between the frequency of a mention and the likelihood that a model will hallucinate the concept assigned to that mention. In other words, LLMs appear equally likely to hallucinate codes for one-off abbreviations, misspellings, and rare wordings as for more frequent spans of text.

\subsection{Entity extraction and hallucinations}
In the Mention and E2E settings, there is the possibility that an LLM will extract a mention for which no SCTID exists, i.e. it may extract spans which should not and cannot be labelled using the target ontology. In the Oracle Span setting, this cannot happen unless the ground truth annotations themselves are incorrect.
A reasonable hypothesis would be that LLMs are more likely to hallucinate codes for these spans, since there is no valid label to assign and since LLMs struggle with refusing to produce output in such cases~\cite{pan2025llmsrefusequestionsknow}.

For all models, we count which mention texts are associated with the highest frequency of hallucinated codes. For the 100 mentions with the highest hallucination rates, we check whether the ground truth data contains any instance where an identical mention was assigned to any concept. If such an instance exists, this implies that there is an SCTID which the model \textit{could} have found to avoid hallucinating; if no such instance exists, it implies that  the given mention may not actually correspond to any SCTID, in which case
hallucination may have been inevitable.
Table~\ref{tab:proportion-solvable-hallucinations} reports how many of these mentions exist in the ground truth data.

Across all models, at least half of the highly-hallucinatory mention texts \textit{do} exist in the ground truth. In other words, hallucinated SCTIDs are not simply a consequence of poor entity extraction.

Table~\ref{tab:high-hallu-mentions} shows the top-5 spans of text which are most likely to be associated with a hallucinated concept ID for each model and setting. 
We observe a remarkable consistency in the set of terms associated with the most hallucinations: a majority are very short abbreviations, many relating to blood panel measurements.
Although the same \textit{terms} are associated with a high degree of hallucination across all models, the hallucinations themselves vary. Claude Opus and Sonnet tend to fabricate codes that resemble an SCTID, or to include SCTIDs which are real but outdated. GPT-5.2 has a stronger tendency to ignore the prompt telling it to produce SNOMED CT codes, and instead produces codes from other vocabularies such as LOINC. These codes are occasionally valid for the text in question, e.g. \texttt{787-2} for \texttt{MCV} with meaning \texttt{MCV [Entitic mean volume] in Red Blood Cells by Automated count}; though just as often they are incorrect, such as \texttt{33728-7} meaning \texttt{Size.maximum dimension in Tumor} assigned to the same text.

\begin{table}[]
    \centering
    \scalebox{0.8}{
    \begin{tabular}{l r}
    \toprule
    \textbf{Model} & \textbf{}\\\midrule
        GPT-5.2 (E2E) & 50\% \\
        GPT-5.2 (Mention) & 75\% \\
        GPT-5.2 (Oracle Span) & 100\% \\
        \midrule
        
        Claude Opus 4.6 (E2E) & 79\% \\
        Claude Opus 4.6 (Mention) & 66\% \\
        Claude Opus 4.6 (Oracle Span) & 100\% \\
        \midrule
        
        Claude Sonnet 4.5 (E2E) & 81\% \\
        Claude Sonnet 4.5 (Mention) & 72\% \\
        Claude Sonnet 4.5 (Oracle Span) & 100\% \\
        
        \bottomrule
    \end{tabular}
    }
    \caption{Of the 100 mentions which receive the most hallucinations, what proportion exist in the ground truth data. A low percentage implies that a model hallucinates primarily in cases where there is no valid code to assign, i.e. when it has no choice but to either hallucinate or refuse to produce output. A higher percentage implies that a model is prone to hallucinate even in cases where there is a valid code which it could output.
    }
    \label{tab:proportion-solvable-hallucinations}
\end{table}

\begin{table*}[]
    \centering
    \scalebox{0.85}{
    \begin{tabular}{l l}
    \toprule
      \textbf{Model}
    & \textbf{Phrases with most hallucinations}
    \\\midrule
    GPT-5.2 (E2E) & WBC, PTT, Plt, Left chest tube$^\dagger$, ciprofloxacin$^\dagger$ \\
    GPT-5.2 (Mention) & WBC, RBC, Plt, MCH, MCV \\
    GPT-5.2 (Oracle Span) & MCV, MCHC, MCH, RDW, BLOOD WBC \\
    \midrule

    Claude Opus 4.6 (E2E) & RDW, MCH, AnGap, HCO3, PTT \\
    Claude Opus 4.6 (Mention) & MCH, RDW, AnGap, TotBili, HCO3 \\
    Claude Opus 4.6 (Oracle Span) & MCHC, MCH, RDW, MCV, AnGap \\
    Claude Opus 4.6 (RAG) & leptomeningeal enhancement$^\dagger$ \\
    \midrule

    Claude Sonnet 4.5 (E2E) & MCHC, MCV, MCH, RDW, AnGap \\
    Claude Sonnet 4.5 (Mention) & Hct, RDW, MCHC, MCH, MCV \\
    Claude Sonnet 4.5 (Oracle Span) & MCH, RDW, MCV, MCHC, AnGap \\
    Claude Sonnet 4.5 (RAG) & foot drop \\
    \bottomrule
        \end{tabular}
        }
    \caption{Phrases which are associated with the greatest likelihood of a hallucinated concept ID. The 5 most-hallucinated phrases are shown for each model, separated by commas. $^\dagger$These phrases are never annotated in the ground truth data.}
    \label{tab:high-hallu-mentions}
\end{table*}

\subsection{SNOMED Extensions}
Across settings, all of the LLM baselines and Amazon Medical Comprehend output a mixture of codes from multiple SNOMED extensions in addition to codes from the ``standard'' International edition.

\begin{table}[]
    \centering
    \scalebox{0.85}{
    \begin{tabular}{l r}
    \toprule
    \multicolumn{2}{l}{\textbf{Model} 
    \hfill\textbf{Deprecated $\downarrow$}}
    \\\midrule
    GPT-5.2 (E2E) 
    & 4.60\% 
    \\
    GPT-5.2 (Mention) 
    & 4.60\% 
    \\
    GPT-5.2 (Oracle Span) 
    & 4.34\% 
    \\
    GPT-5.2 (RAG) 
    & 0.57\% 
    \\
    \midrule

    Claude Opus 4.6 (E2E) 
    & 4.32\% 
    \\
    Claude Opus 4.6 (Mention) 
    & 4.70\% 
    \\
    Claude Opus 4.6 (Oracle Span) 
    & 4.50\% 
    \\
    Claude Opus 4.6 (RAG) 
    & 0.63\% 
    \\
    \midrule
    
    Claude Sonnet 4.5 (E2E) 
    & 3.61\%
    \\
    Claude Sonnet 4.5 (Mention) 
    & 4.31\% 
    \\
    Claude Sonnet 4.5 (Oracle Span) 
    & 3.95\%
    \\
    Claude Sonnet 4.5 (RAG) 
    & 0.59\% 
    \\
    \midrule

    Amazon Comprehend Medical
    & 3.47\%
    \\\bottomrule
    \end{tabular}
    }
    \caption{Macro-averaged proportion of outputs representing real but inactive SCTIDs.}
    \label{tab:inactive-codes}
\end{table}

\begin{table}[]
    \centering
    \scalebox{0.85}{
    \begin{tabular}{l r}
    \toprule
    \multicolumn{2}{l}{\textbf{Model} 
    \hfill\textbf{Invalid $\downarrow$}}
    \\\midrule
    GPT-5.2 (E2E) 
    & 18.72\% 
    \\
    GPT-5.2 (Mention) 
    & 27.71\% 
    \\
    GPT-5.2 (Oracle Span) 
    & 36.88\% 
    \\
    GPT-5.2 (RAG) 
    & 0.00\% 
    \\
    \midrule

    Claude Opus 4.6 (E2E) 
    & 41.60\%
    \\
    Claude Opus 4.6 (Mention) 
    & 24.80\% 
    \\
    Claude Opus 4.6 (Oracle Span) 
    & 34.07\% 
    \\
    Claude Opus 4.6 (RAG) 
    & 0.02\% 
    \\    
    \midrule
    
    Claude Sonnet 4.5 (E2E) 
    & 54.46\% 
    \\
    Claude Sonnet 4.5 (Mention) 
    & 34.07\%  
    \\
    Claude Sonnet 4.5 (Oracle Span)
    & 51.49\%
    \\
    Claude Sonnet 4.5 (RAG) 
    & 0.02\%  
    \\
    \midrule
    
    Amazon Comprehend Medical
    & 0.00\%
    \\\bottomrule
    \end{tabular}
    }
    \caption{Macro-averaged proportion of outputs representing invalid SCTIDs.}
    \label{tab:invalid-codes}
\end{table}

GPT-5.2 includes codes from 5, 7, and 8 different extensions in the End-to-End, Mention, and Oracle Span settings respectively; Opus includes 7, 5, and 2; and Sonnet includes 10, 6, and 5. (This ignores invalid extensions such as \texttt{3000119}, which does not exist in the SNOMED CT Namespace Identifier Registry; across all models we observe roughly a dozen such nonexistent extensions in total.) AWS Comprehend outputs codes from 14 distinct extensions spanning organizations from four continents, although all of these are real extensions rather than invented ones.

Looking at the Oracle Span setting, where Opus includes just 2 distinct extensions, we find that these come from namespace \texttt{1000000}, belonging to the United Kingdom's National Health Service, and namespace \texttt{1000119}, belonging to Kaiser Permanente, an American corporation. These extensions therefore reflect vastly different healthcare systems, national versus corporate levels of organization, and consequently different needs and motivations underlying the choice of concepts included in their respective extensions.
Thus, while these codes are not incorrect \textit{per se}, they nevertheless represent a degree of inconsistency which may negatively impact the usefulness of these model outputs in a healthcare setting. This inconsistency will be all the greater in settings which mix codes from a larger number of extensions.

\subsection{Deprecated Codes}
Table~\ref{tab:inactive-codes} reports the proportion of codes from each model which are listed as ``inactive'' in the United States edition 2025-03-01 release of SNOMED.

Between 3\% and 5\% of codes are deprecated on average, except in RAG settings where less than 1\% of codes are deprecated.
The proportion of deprecated codes is otherwise roughly stable across model providers and experiment settings. 

These numbers highlight LLMs' limited ability to handle versioned data. The continual nature of ontology updates, combined with the fact that concept IDs may occur in training data with no indication of which version they are associated with, make it challenging to consistently eliminate old codes and incorporate current ones into these models.

\subsection{Concept Depth}
\begin{figure}
    \centering
    \includegraphics[width=\linewidth]{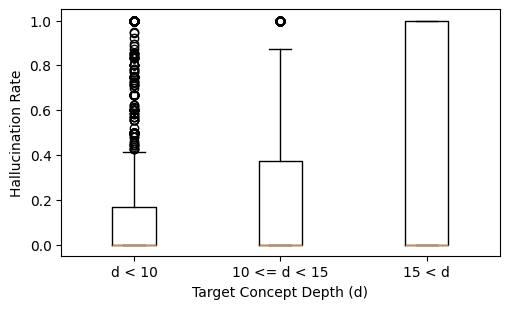}
    \caption{Hallucination rate vs target concept depth. Concepts which are deeper in the SNOMED CT hierarchy are associated with a higher risk of hallucination.}
    \label{fig:boxplot}
\end{figure}

By taking a concept's depth in the SNOMED CT hierarchy as a proxy for its specificity, we find that LLMs are more likely to hallucinate codes for more specific concepts.
As illustrated in Figure~\ref{fig:boxplot}, we observe a highly statistically significant interaction between the depth of a ground truth concept label in the SNOMED CT hierarchy and the likelihood that a model will hallucinate its output for the corresponding span, though the effect size is small ($\rho = 0.09$, $p = 3\times10^{-22}$ averaged across all models). 
This likely reflects the fact that more general codes are better represented in the available training data, and therefore more likely to have been seen by models.
The effect is strongest for Claude Sonnet ($\rho = 0.12$, $p = 1.5\times10^{-14}$); Sonnet is also the smallest model included in our experiments, so it is perhaps not surprising that it should exhibit the poorest recall for more specialized concepts. 

For a discussion of concept depth as it relates to \textit{non}-hallucinated model outputs, see Appendix~\ref{sec:appendix-concept-depth}.

\subsection{Invalid Codes}
Without RAG, between 18\% and 54\% (Table~\ref{tab:inactive-codes}) of the SCTIDs produced by LLMs are \textit{invalid}; that is, beyond simply not existing in any current release of SNOMED CT, they could never exist as they violate one or more of the basic formatting requirements which SCTIDs are required to follow.

The overwhelming majority of these violations are due to invalid check digits. An SCTID must end with a single-digit checksum, which is incorrect in more than 99\% of hallucinated codes in most settings.
There are only two settings in which other error types make up a significant proportion of invalid codes. In the GPT-5.2 Oracle Span setting, roughly 4\% of hallucinations include an invalid partition identifier, which indicates the type of component the SCTID refers to (concept, relationship, etc). In the GPT-5.2 Mention setting, 27\% of hallucinations include an invalid partition identifier.

\section{Related Work}

Recent work has established robust frameworks for evaluating biomedical entity linking. \citet{kartchner-etal-2023-comprehensive} benchmark nine models across dimensions including scalability and zero-shot robustness, while BioEL \citep{bathala-etal-2025-bioel} standardizes training across ontologies like UMLS \citep{umls} and MeSH \citep{mesh}. However, these works primarily target biomedicine and traditional discriminative architectures, leaving the specific performance trade-offs and hallucination risks of LLMs in clinical settings largely unexplored.

\citealt{ChangSung2024-snomed-in-llm} examine how SNOMED CT has been integrated into prior LLM research. They find that most research uses LLMs to produce contextual embedding vectors for entities, requiring fine-tuning on some downstream task. A majority of research integrates SNOMED CT in pretraining or fine-tuning, either lexically or in graph form.

\begin{figure}
    \centering
    \includegraphics[width=\linewidth]{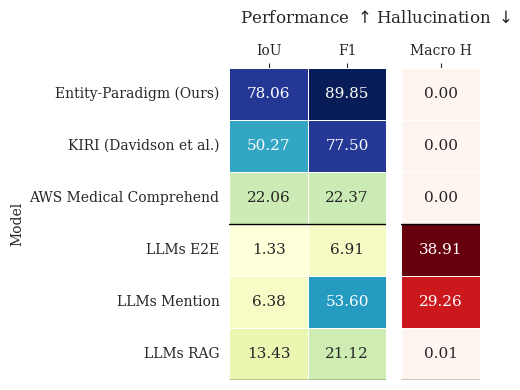}
    \caption{Summary of results grouped by setting, demonstrating high average hallucination rate and low average accuracy for generalist LLMs compared to task-specific models.}
    \label{fig:summary}
\end{figure}

\citealt{vishwanath2025generalistlargelanguagemodels} compare generalist frontier models against specialized clinical AI tools like OpenEvidence\footnote{\url{https://www.openevidence.com/}} and UpToDate Expert AI\footnote{\url{https://www.wolterskluwer.com/en/solutions/uptodate}} on MedQA (\citealt{jin2020disease}; USMLE-style questions) and HealthBench (\citealt{arora2025healthbenchevaluatinglargelanguage}; clinical judgment). The results indicate that generalist LLMs outperform specialized tools on clinical judgment tasks, challenging marketing claims and emphasizing the need for transparent, independent evaluation before deployment of patient-facing workflows.

TrialMind, a pipeline built on GPT-4, was bench-marked against 100 published reviews involving over 2,220 studies \cite{wang2025accelerating}.
Medical experts preferred TrialMind’s summaries over GPT-4, highlighting that the best clinical outcomes are achieved through a ``hybrid'' LLM-human collaboration \cite{wang2025accelerating}.

Me-LLaMA, developed through continual pretraining and task-specific instruction tuning, demonstrated performance comparable to GPT-4 in diagnosing complex clinical cases \cite{xie2025medical}.
GPT-4 often outperform Google’s Med-PaLM 2 \cite{singhal2025toward} in terms of both hallucination rate and human-centric measures like empathy and readability \cite{zeba2025mitigatinghallucinationshealthcarellms}.

Most studies investigating the performance of LLMs for autonomous medical coding \cite{Soroush-poor-medical,kwan2024largelanguagemodelsgood,hou2025enhancing,yuan-etal-2025-toward,motzfeldt-etal-2025-code} focus on the task of clinical notes to medical billing codes using ICD dataset. While LLMs have been shown to generally struggle with autonomous medical coding \cite{Soroush-poor-medical}, \citealt{kwan2024largelanguagemodelsgood} propose a two-stage retrieve/rerank framework to improve performance. However, the restrictions on the input format render it incompatible with real world clinical notes.
\citealt{yuan-etal-2025-toward} propose adaptations to enhance the performance of LLMs 
and fix errors from hierarchical misalignments. 
 In contrast, the focus of this paper is SNOMED coding. While ICD-9 and ICD-10 coding involve document level information and consolidation, SNOMED coding is more granular and ties medical concepts directly to spans of text.

\section{Conclusion}

We have demonstrated that medical entity linking against a controlled ontology provides good testbed for distinguishing hallucinations from other kinds of erroneous LLM output. We show that at least 20\%, and in some cases nearly 55\% of the SNOMED IDs (SCTIDs) produced by frontier LLMs on an entity linking task do not exist in the ontology these models were instructed to use. Of these erroneous codes, roughly one-third are fabrications which do not satisfy the formatting and other validation requirements to be interpreted as valid SCTIDs and are therefore highly unlikely to have been seen during training. Roughly 4.5\% of erroneous codes \textit{were} valid at some point in time, but have since been deprecated, highlighting that these models have difficulty keeping track of differences between ontology versions and are prone to conflating them. All models demonstrate a similar tendency to conflate the core SNOMED CT ontology with SNOMED CT \textit{extensions} produced by national or corporate healthcare organizations. Since these extension codes \textit{are} technically valid SCTIDs, they represent a particularly subtle form of error that may be difficult to catch if one is not specifically looking for it. Overall, these results demonstrate that LLMs still lack the reliability required for high-stakes medical applications, and highlight areas of particular concern for prospective users of LLM-based medical coding systems. As a future direction, we believe Model Context Protocol (MCP) tools\footnote{\url{https://www.anthropic.com/news/model-context-protocol}} which are specialized for entity linking offer a better way to align LLM outputs with medical knowledge \cite{elsayed2025mcpaiprotocoldrivenintelligenceframework} and to assist with medical coding tasks. 

\section{Bibliographical References}\label{sec:reference}

\bibliographystyle{lrec2026-natbib}
\bibliography{bib}


\clearpage

\newpage
\appendix

\section{Reannotation methodology}\label{sec:datacleanup}
As noted in Section~\ref{sec:data}, the original SNOMED CT entity linking challenge~\cite{snomedct-challenge} suffers from inconsistent labelling, such that two identical phrases in different texts may be annotated with a concept link in one case but left unannotated in the other. 
This provides an inconsistent training signal for models which require training data, and imposes an upper limit on the accuracy a model can achieve on the test set, as it comes down to random chance whether a given semantically-correct concept link will have been included among the ground truth spans. For this reason, the upper limit of scores achievable on the original test set is likely much lower than the theoretical limit of 1.0 IoU, since past a certain point, improving a model's score would require knowledge of which spans were actually covered by the ground truth annotations so that the model could avoid producing semantically-correct outputs which happen to have been missed by the annotators.\footnote{It is also for this reason that we report the score of our \texttt{entity-paradigm} system on the test set, even though it is the model used to prepare the test set and is therefore unfairly advantaged. Our score represents an approximate upper bound on the scores that should be achievable on this dataset when a model's level of granularity, phrase segmentation choices, set of known concepts, etc are perfectly matched to the test set.}

Furthermore, the challenge utilized a simplified subset of SNOMED CT (May 2023 International Edition) focusing only on three sub-hierarchies (findings, procedures, and body structures), which limits the breadth of model behaviours it can be used to evaluate.

These factors motivated our decision to reannotate the dataset for this work. We use an automated system to assign concept links, which ensures a greater degree of consistency across documents and eliminates the possibility of inter-annotator disagreement. The system used to assign concept links is trained on a corpus of clinical notes which have been manually annotated by clinicians: its training inputs comprise 1,137,651 manual annotations covering 380,588 SCTIDs.
We employ a broader set of semantic types than the original dataset in order to capture a more comprehensive range of clinical entities. The semantic types used in our evaluation were: disorder, procedure, finding, morphologic abnormality, body structure, observable entity, and regime/therapy. 
This expanded scope allows for a more thorough assessment of models' ability to handle diverse medical terminology across the full clinical spectrum, including pathological states, therapeutic interventions, clinical observations, anatomical components, and treatment regimens.
Medicinal products and substances (e.g., ciprofloxacin) are excluded from this evaluation, as they fall outside the semantic types used.

Figure~\ref{fig:reannotation-examples} shows a sample of sentences where the set of annotations has changed as a result of our reannotation.

Since concept links were assigned automatically, we add an LLM filtering step to flag and remove links that may have been assigned in error. In total, Claude Haiku 4.5 flags {13\%} of the automatically applied concept links for removal.  We present a random sample of the removed links in Table~\ref{tab:flagged-rows}. We note that while the LLM filtering does remove some legitimate mistakes (e.g. ``he admitted'' labelled as a hospital admission), it can also be overzealous about removing valid links when they occur in a negated context (``no vomiting'') or when the concept wording does not match its expectations (insisting that ``VII'' is a cranial nerve, not a facial nerve, when in fact it is both).

\begin{table*}
    \begin{tabular}{p{6.1in}}
    \toprule
    \textbf{Original:} There exists some crowded vasculature in the \\$\underset{\color{blue}\textsf{41224006 | Structure of lower lobe of left lung (body structure) |}}{\hspace{-13em}\underline{\textsf{left lower lobe}}}\hspace{-13em}$ area in retrocardiac position possibly suggesting some $\underset{\color{blue}\textsf{46621007 | Atelectasis (disorder) |}}{\hspace{-5.5em}\underline{\textsf{atelectasis}}}\hspace{-5.5em}$ but acute discrete parenchymal $\underset{\color{blue}\textsf{231287002 | Infiltration (procedure) |}}{\hspace{-7em}\underline{\textsf{infiltrates}}}\hspace{-5em}$ identifying a $\underset{\color{blue}\textsf{233604007 | Pneumonia (disorder) |}}{\hspace{-6em}\underline{\textsf{pneumonia}}}\hspace{-6em}$ cannot be seen.\\[1em]
    \midrule
    \textbf{Updated:} There exists some crowded $\underset{\color{blue}\textsf{59820001 | Blood vessel structure (body structure) |}}{\hspace{-10.5em}\colorbox{LightYellow}{\underline{\textsf{vasculature}}}}\hspace{-10.5em}$ in the \\$\underset{\color{blue}\textsf{41224006 | Structure of lower lobe of left lung (body structure) |}}{\hspace{-13em}\underline{\textsf{left lower lobe}}}\hspace{-13em}$ area in retrocardiac position possibly suggesting some $\underset{\color{blue}\textsf{46621007 | Atelectasis (disorder) |}}{\hspace{-5.5em}\underline{\textsf{atelectasis}}}\hspace{-5.5em}$ but acute discrete $\underset{\color{blue}\textsf{409609008 | Radiologic infiltrate of lung (disorder) |}}{\hspace{-5.5em}\colorbox{LightYellow}{\underline{\textsf{parenchymal infiltrates}}}}\hspace{-5.5em}$ identifying a $\underset{\color{blue}\textsf{233604007 | Pneumonia (disorder) |}}{\hspace{-6em}\underline{\textsf{pneumonia}}}\hspace{-6em}$ cannot be seen.\\[1em]
    \hline
    \midrule
    \textbf{Original:} BLOOD $\underset{\color{blue}\textsf{33747003 | Glucose measurement, blood (procedure) |}}{\hspace{-12.5em}\underline{\textsf{Glucose}}}\hspace{-12.5em}$-219*\\
    $\underset{\color{blue}\textsf{105011006 | Blood urea nitrogen measurement (procedure) |}}{\hspace{-15em}\underline{\textsf{UreaN}}}\hspace{-15em}$-19 \\
    $\underset{\color{blue}\textsf{70901006 | Creatinine measurement (procedure) |}}{\hspace{-12em}\underline{\textsf{Creat}}}\hspace{-12em}$-0.9 \\
    $\underset{\color{blue}\textsf{312469006 | Blood sodium measurement (procedure) |}}{\hspace{-14.5em}\underline{\textsf{Na}}}\hspace{-14.5em}$-138 \\
    \ K-4.3 \\
    $\underset{\color{blue}\textsf{104589004 | Chloride measurement, blood (procedure) |}}{\hspace{-15em}\underline{\textsf{Cl}}}\hspace{-15em}$-104 \\
    $\underset{\color{blue}\textsf{312471006 | Blood bicarbonate measurement (procedure) |}}{\hspace{-14em}\underline{\textsf{HCO3}}}\hspace{-14em}$-26 \\
    $\underset{\color{blue}\textsf{25469001 | Anion gap measurement (procedure) |}}{\hspace{-11.5em}\underline{\textsf{AnGap}}}\hspace{-11.5em}$-12\\[0.5em]
    \midrule
    \textbf{Updated:} $\underset{\color{blue}\textsf{33747003 | Glucose measurement, blood (procedure) |}}{\hspace{-9em}\colorbox{LightYellow}{\underline{\textsf{BLOOD Glucose}}}}\hspace{-9em}$-219* \\
    $\underset{\color{blue}\textsf{105011006 | Blood urea nitrogen measurement (procedure) |}}{\hspace{-15em}\underline{\textsf{UreaN}}}\hspace{-15em}$-19 \\
    $\underset{\color{blue}\textsf{113075003 | Creatinine measurement, serum (procedure) |}}{\hspace{-14.5em}\underline{\textsf{Creat}}}\hspace{-14.5em}$-0.9 \\
    $\underset{\color{blue}\textsf{104934005 | Sodium measurement, serum (procedure) |}}{\hspace{-15em}\colorbox{LightYellow}{\underline{\textsf{Na}}}}\hspace{-15em}$-138 \\
    $\underset{\color{blue}\textsf{271236005 | Serum potassium measurement (procedure) |}}{\hspace{-16em}\colorbox{LightYellow}{\underline{\textsf{K}}}}\hspace{-16em}$-4.3 \\
    $\underset{\color{blue}\textsf{271238006 | Serum chloride measurement (procedure) |}}{\hspace{-15em}\colorbox{LightYellow}{\underline{\textsf{Cl-}}}}\hspace{-15em}$104 \\
    $\underset{\color{blue}\textsf{312471006 | Blood bicarbonate measurement (procedure) |}}{\hspace{-14em}\colorbox{LightYellow}{\underline{\textsf{HCO3-}}}}\hspace{-14em}$26 \\
    $\underset{\color{blue}\textsf{271057005 | Serum anion gap measurement (procedure) |}}{\hspace{-14em}\colorbox{LightYellow}{\underline{\textsf{AnGap}}}}\hspace{-14em}$-12\\[1em]
    \bottomrule
    \end{tabular}
    \caption{Sample of differences resulting from \colorbox{LightYellow}{reannotation}. Our update provides broader coverage by adding new links for ``vasculature'' and ``K''. We replace an incorrect mapping for ``infiltrates'' with a more correct, and longer, mapping for ``parenchymal infiltrates''. Our concept assignments are more internally consistent: blood measurements are labelled as such in our output, where the original occasionally uses a more generic measurement concept.}
    \label{fig:reannotation-examples}
\end{table*}

\begin{table*}\centering
    \scalebox{.74}{
    \begin{tabular}{lp{4cm}p{5cm}p{9cm}}
        \toprule
          \textbf{Concept ID}
        & \textbf{FSN}
        & \textbf{Context}
        & \textbf{Justification}
        \\\midrule
        
        119265000 & Assisting (procedure) & Her daughter [...] was able to \textbf{assist} with much of the history & Assist means support/help, not a medical procedure. Context is supportive role, not clinical intervention. \\
        56052001 & Facial nerve structure (body structure) & V, \textbf{VII}: Facial strength and sensation intact and symmetric & VII is cranial nerve designation, not facial nerve body structure. \\
        32485007 & Hospital admission (procedure) & He \textbf{admitted} that he had lost 10 lbs & "admitted" used as past tense verb describing weight loss acknowledgment, not hospital admission procedure. \\
        420227002 & Recommendation to (procedure) & per palliative \textbf{recommendations} will be discharged home & Concept too specific; "recommendations" refers to discharge plans, not procedure recommendations. \\
        58000006 & Patient discharge (procedure) & \textbf{DISCHARGE} LABS: & DISCHARGE in section heading refers to test type, not patient discharge procedure. \\
        74262004 & Oral cavity structure (body structure) & Lansoprazole \textbf{Oral} Disintegrating Tab 15 mg PO DAILY & Route of administration mismatched with anatomical structure concept. \\
        31156008 & Structure of left half of body (body structure) & you had pain on your \textbf{left} back and under your ribs & Concept 'left' is overly broad; matching text specifies 'left back' only. \\
        307818003 & Weight monitoring (regime/therapy) & Remember to \textbf{weigh yourself every morning} & FSN is literal instruction, not therapeutic procedure concept. Misaligned with clinical intent. \\
        422400008 & Vomiting (disorder) & Her headaches ia also associated with nausea and dry heaves, no \textbf{vomiting}. & Explicit negation: "no vomiting" contradicts vomiting concept despite guideline allowing negation. 
        \\\bottomrule
    \end{tabular}
    }
    \caption{Random sample of annotations which were removed by LLM filtering, and justification provided by LLM for removal.}
    \label{tab:flagged-rows}
\end{table*}

\section{Embedding-Weighted IoU}\label{sec:ewiou}
Standard entity linking metrics such as IoU and F1 conflate a model's ability to produce semantically-valid concept links with its ability to follow a specific annotation style. A model which annotates \texttt{left and right} as the concept \texttt{Right and left (qualifier value)} is as correct, in semantic terms, as one which labels \texttt{left} as \texttt{Left (qualifier value)} and \texttt{right} as \texttt{Right (qualifier value)}. However, in terms of IoU or F1, segmenting the phrase in this way incurs the same penalty as if the model had mislabelled it entirely.

We propose an embedding-weighted IoU score (EWIoU) which awards partial points based on the cosine similarity between dense representations of the observed and expected concepts as a means of decoupling semantic correctness from segmentation or specificity.

If $P_{\mathrm{c}}$ is the set of characters which were predicted to belong to some concept $c$, and $G_{\mathrm{c}}$ is the set of characters labelled with  concept $c$ in the ground truth data, IoU computes the following quantity:

\[
    \mathrm{IoU}_{c} = \frac{|P_{c} \cap G_{c}|}{|P_{c} \cup G_{c}|}
\]

Now let $\mathbf{e}_c$ be a dense embedding representing concept $c$, and likewise let $\mathbf{e}_\mathrm{char}$ be the embedding for the concept assigned to the character $\mathrm{char}$ by some model. (We use the same OpenAI \texttt{text-embedding-3-large} embeddings as in the RAG retriever.) Then the embedding-weighted IoU is:

\[
    \mathrm{EWIoU}_{\textrm{c}} = 
    \frac{
        \sum_{\mathrm{char} \in P_{c} \cup G_{c}} \mathbf{e}_{c} \cdot \mathbf{e}_{\mathrm{char}}
    }{
        |P_{c} \cup G_{c}|
    }  
\]

\begin{table*}
    \centering
    \begin{tabular}{l ll}
        \toprule
        
        \multicolumn{3}{c}{
        \textbf{Task-Specific Models}
        }\\\midrule

          \textbf{Model} 
        & \textbf{IoU $\uparrow$} 
        & \textbf{EWIoU $\uparrow$} 
        \\\midrule
        
        entity-paradigm (Ours)
        & \color{lightgray}\dashuline{78.06\%} 
        & \color{lightgray}\dashuline{78.06\%}\\\midrule
                
        KIRI
        & 50.49\% 
        & 63.93\%\\

        Amazon Comprehend Medical 
        & 22.06\%
         & 41.68\%\\

        \midrule
        
        \multicolumn{3}{c}{
        \textbf{General-Purpose Frontier LLMs}}\\\midrule

          \textbf{Model} 
        & \textbf{IoU $\uparrow$} 
        & \textbf{EWIoU $\uparrow$} 
        \\\midrule
        
        GPT-5.2 (E2E) 
        & 0.15\% 
         & 2.93\%\\

        Claude Opus 4.6 (E2E) 
        & 2.23\%
         & 12.66\%\\

        Claude Sonnet 4.5 (E2E) 
        & 1.60\%
         & 10.52\%\\
         \midrule

        GPT-5.2 (Mention) 
        & 5.39\% 
         & 25.47\%\\

        Claude Opus 4.6 (Mention) 
        & 8.98\% 
         & 27.34\%\\

        Claude Sonnet 4.5 (Mention) 
        & 4.78\% 
         & 17.05\%\\
         \midrule

        GPT 5.2 (RAG)
        & 11.66\%
         & 41.14\%\\

        Claude Opus 4.6 (RAG)
        & \textbf{14.56\%}
         & \textbf{45.39\%}\\

        Claude Sonnet 4.5 (RAG)
        & 14.08\%
         & 43.12\%\\
        \midrule
        
        \multicolumn{3}{c}{
        \textbf{Oracle Setting}}\\\midrule

          \textbf{Model} 
        & \textbf{IoU $\uparrow$} 
        & \textbf{EWIoU $\uparrow$} 
        \\\midrule
        entity-paradigm (Ours; Oracle Span)
        & \color{lightgray}84.45\%
         & \color{lightgray}84.45\%\\
         \midrule
        
        GPT-5.2 (Oracle Span)
        & 6.69\%
         & 38.83\%\\
         
        Claude Opus 4.6 (Oracle Span)
        & \textbf{12.36\%}
         & \textbf{40.43\%}\\

        Claude Sonnet 4.5 (Oracle Span)
        & 8.02\%
         & 29.67\%\\
         \bottomrule
    \end{tabular}
    \caption{Embedding-Weighted IoU scores for all models.}
    \label{tab:ewiou}
\end{table*}

Table~\ref{tab:ewiou} compares the plain and embedding-weighted IoU scores for all models in all settings.
Note that the EWIoU score for \texttt{entity-paradigm} is identical to its IoU score. This is expected, as it is the model that was used to update the data set with more comprehensive concept links, and it should trivially reproduce its own choices regarding phrase segmentation and level of detail. Thus it does not benefit from a more relaxed evaluation.

Other models see significant improvements, however: among the task-specific models, KIRI gains +13 percentage points and Amazon Comprehend Medical gains +19. For the generalist LLMs, the largest absolute gains are seen in the RAG setting, where all models gain upwards of +30 points; the absolute gain is more modest in the other settings, but is still quite large in relative terms, representing nearly a twenty-fold improvement over regular IoU in the worst-performing setting.

This discrepancy highlights that all models produce a relatively large number of outputs which are semantically close to the target concept even when they do not match the target exactly.
This demonstrates a shortcoming of exact-match metrics like IoU in settings where multiple annotations are potentially valid, and where correctness may depend more on specific annotation conventions than on the underlying semantics of the concept link.
We suggest that fuzzy evaluations like EWIoU may give a more accurate view of semantic accuracy in such settings.

\section{Additional Remarks on\\Concept Depth}\label{sec:appendix-concept-depth}
\begin{figure}
    \centering
    \includesvg[inkscapelatex=false,width=\linewidth]{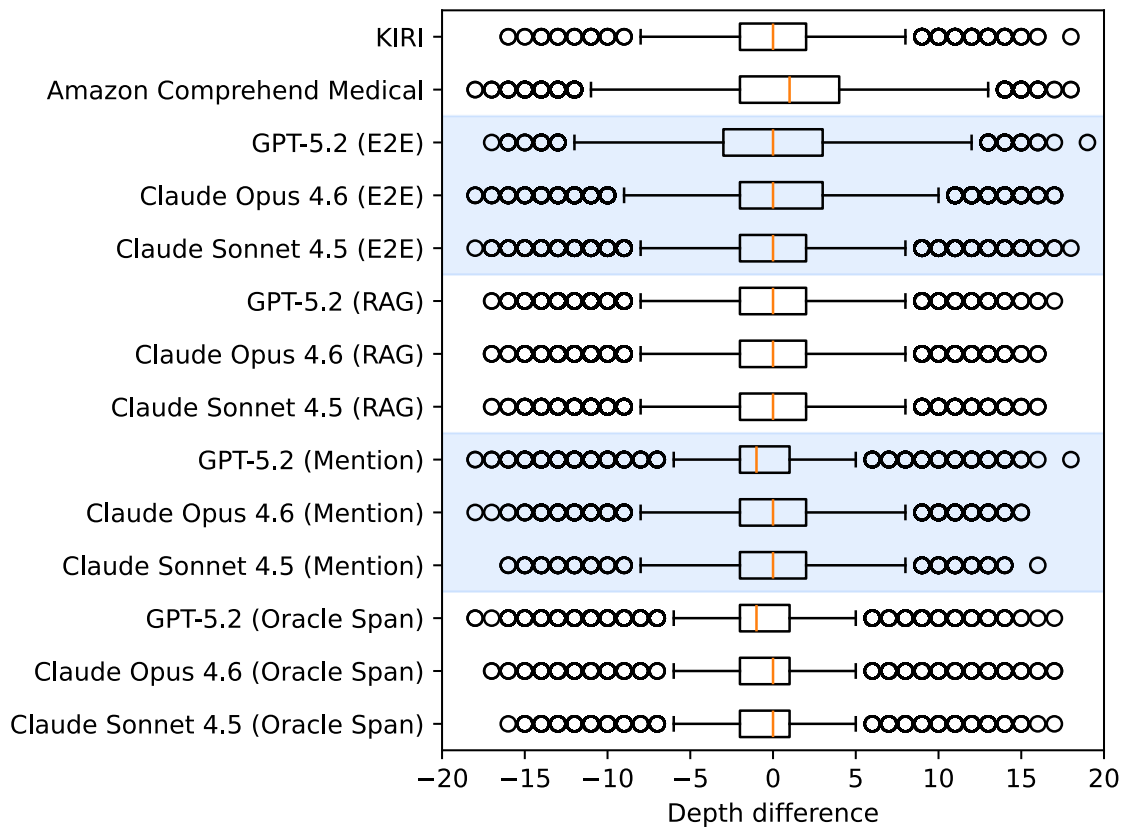}
    \caption{Difference between depth of predicted and target concepts in the SNOMED CT hierarchy.}
    \label{fig:depthdiff}
\end{figure}

When a model produces a real, but incorrect, concept, it is roughly equally likely to be deeper in the hierarchy (i.e. more specific) than the target as it is to be shallower (i.e. more generic; Figure~\ref{fig:depthdiff}). Across all models, the median difference in depth between observed and expected concepts is zero, with the exception of Amazon Comprehend Medical, where the predicted concept is one level deeper than the target on average, and GPT-5.2 (Mention and Oracle Span) where it is one level shallower.

\begin{figure}
    \centering
    \includesvg[width=\linewidth]{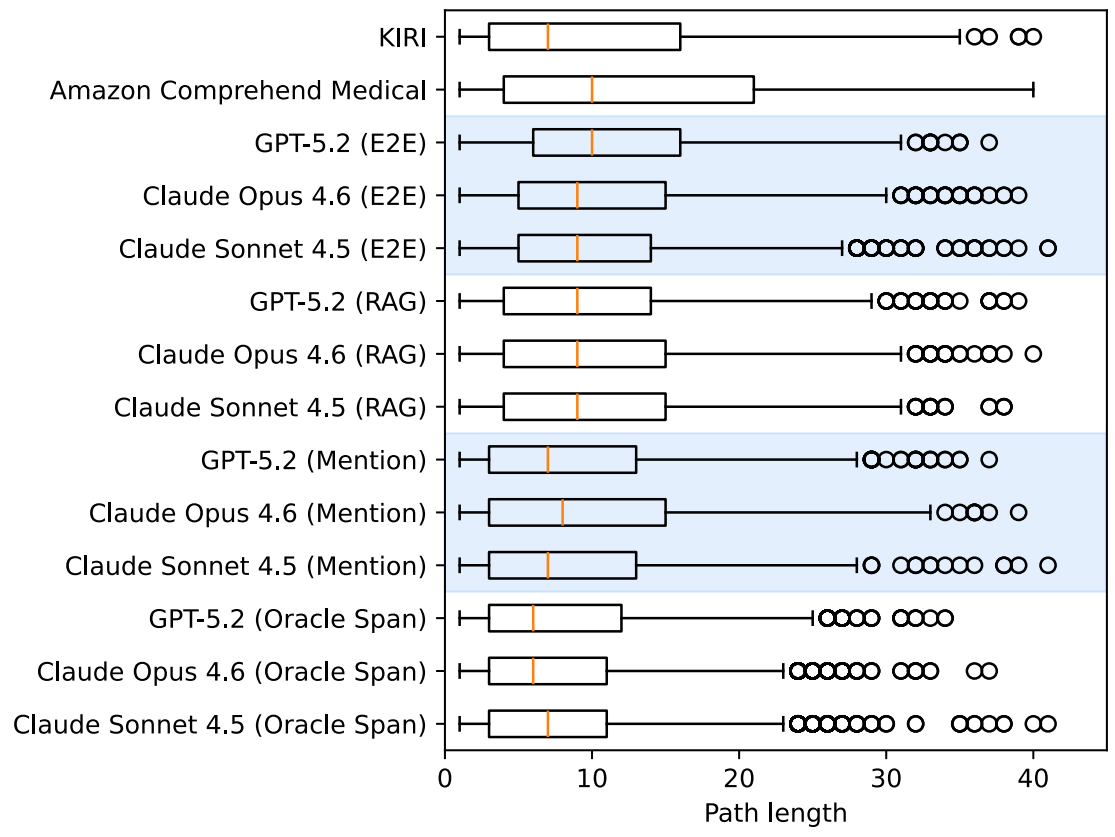}
    \caption{Length of shortest path from predicted to target concept in the SNOMED CT hierarchy.}
    \label{fig:pathlength}
\end{figure}

To further investigate the nature of non-hallucinated errors, we compute the length of the path linking an erroneous prediction to the corresponding target in the SNOMED CT hierarchy. We walk up the hierarchy from the predicted concept to the lowest common ancestor of the target and the prediction, then back down to the target. A boxplot of the resulting path lengths is shown in Figure~\ref{fig:pathlength}.

The shortest mean path lengths are found in the Oracle Span setting. This is unsurprising, as knowledge of the target span also provides information about the expected level of granularity of the target concept.
The longest path lengths are incurred by Amazon Comprehend Medical. Where the other models will label ambiguous mentions (e.g. ``disease'') with accurate but generic concepts (\texttt{Disease (disorder)}), the Amazon model attempts to resolve these to the specific underlying disorder based on context clues. Ignoring the accuracy of this disambiguation, which is outside the scope of the present work, this approach produces concepts at a much different level of granularity than any of the other approaches, which is reflected by the longer path lengths.

\section{Prompt Engineering and Optimization Protocol}
To ensure a robust comparison between generalist LLMs and the task-specific commercial benchmark, we employed a systematic optimization process for all prompt-based models. Rather than relying on a single static prompt, we treated prompt design as a critical experimental variable:

\begin{enumerate}
    \item \textbf{Iterative Optimization}: We conducted an iterative search for the optimal prompt structure using a small, held-out validation set. This included testing various instruction formats, persona assignments (e.g, "You are a medical terminology expert .."), and the inclusion of explicit logical constraints.
    \item \textbf{Reasoning Strategies}: To ensure the generalist models were evaluated at their best capability, we tested several reasoning configurations, including chain-of-thought (CoT). Our optimization process showed that for this task, the inclusion of intermediate reasoning steps or extended "thinking" modes introduced unnecessary variance and lower overall accuracy. Therefore, we utilized optimized direct-response prompts, which provided the most robust and competitive results for the LLMs.
\end{enumerate}
Our \texttt{entity-paradigm} model, the commercial offering by \textit{emtelligent}, used in this study is a proprietary, task-specific architecture that does not rely on prompting. 
It is included as a 'Performance Oracle' to provide a high-water mark for the task.
By ensuring that the generalist LLM prompts were fully optimized, we can more accurately quantify the remaining performance gap between general-purpose models and specialized, state-of-the-art solutions.

\begin{figure*}[h]
\centering
\begin{tcblisting}{
    colback=gray!10,
    colframe=black,
    title=Oracle Span Prompt Template,
    width=0.95\textwidth, % Use almost full width
    listing only,         % Tells latex this is verbatim content
    listing options={
        basicstyle=\ttfamily\footnotesize, % Smaller monospaced font
        breaklines=true,       % AUTOMATICALLY WRAPS LONG LINES
        breakatwhitespace=false,
        columns=fullflexible,  % Better spacing
        keepspaces=true
    }
}

You are a medical terminology expert specializing in SNOMED CT (Systematized Nomenclature of Medicine - Clinical Terms).

Given a clinical note and a list of medical entity mentions extracted from it, your task is to provide the SNOMED CT concept ID for each entity mention.

<clinical_note>
{{ note_text }}
</clinical_note>

## Entity Mentions to Link
{% for entity in entities %}
- ID {{ loop.index0 }}: "{{ entity.mention_text }}" (position {{ entity.start }}-{{ entity.end }})
{% endfor %}

## Instructions
For each entity mention above, provide the most appropriate SNOMED CT concept ID.

Reason about the correct SNOMED CT concept for each entity  in <scratchpad> tags BEFORE you start writing JSON. Your <scratchpad> must be short and concise, in very few words. Once you begin the JSON array, do NOT interrupt it with any commentary, reasoning, or corrections - output the complete array in one unbroken block inside the <json> tags.

STRICT FORMAT RULES:
1. The response MUST contain exactly one JSON array inside the <json> tags
2. The array MUST contain exactly {{ entities|length }} objects, one per entity
3. Every object MUST have exactly two keys: "entity\_id" (integer) and "concept\_id" (string)
4. Do NOT omit "entity_id" from any object. Every single object MUST include both "entity_id" AND "concept_id"
5. The "entity_id" values MUST be {{ range(entities|length)|list|join(', ') }} in that exact order
6. The "concept_id" MUST be a numeric SNOMED CT code as a string (e.g. "89187006")

Example for 3 entities:
<json>
[{"entity_id": 0, "concept_id": "89187006"}, {"entity_id": 1, "concept_id": "38341003"}, {"entity_id": 2, "concept_id": "73211009"}]
</json>

Now output the JSON array for all {{ entities|length }} entities.
\end{tcblisting}
\caption{Prompt template for the \textbf{Oracle Span} setting.}
\label{fig:prompt1}
\end{figure*}

\begin{figure*}[h]
\centering
\begin{tcblisting}{
    colback=gray!10,
    colframe=black,
    title=End-to-End Prompt Template,
    width=0.95\textwidth, % Use almost full width
    listing only,         % Tells latex this is verbatim content
    listing options={
        basicstyle=\ttfamily\footnotesize, % Smaller monospaced font
        breaklines=true,       % AUTOMATICALLY WRAPS LONG LINES
        breakatwhitespace=false,
        columns=fullflexible,  % Better spacing
        keepspaces=true
    }
}

You are a medical terminology expert specializing in SNOMED CT (Systematized Nomenclature of Medicine - Clinical Terms).

Given a clinical note, your task is to:
1. Identify all medical entities (anatomical structures, clinical findings, procedures, substances, disorders, etc.)
2. For each entity, provide its exact character positions in the text and the corresponding SNOMED CT concept ID.

<clinical_note>
{{ note_text }}
</clinical_note>

## Instructions
Extract every medical entity from the clinical note and link each to its SNOMED CT concept ID.

Reason about the correct mention and it's SNOMED CT concept inside <scratchpad> tags BEFORE you start writing JSON. Your <scratchpad> must be short and concise, in very few words. Once you begin the JSON array, do NOT interrupt it with any commentary, reasoning, or corrections - output the complete array in one unbroken block inside the <json> tags.

Respond with ONLY a JSON array, no other text. Each element must have:
- "text": the exact entity mention as it appears in the note - must be present
- "start": INT character offset where the mention begins (0-indexed from the start of the note text) - must be present
- "end": INT character offset where the mention ends (exclusive) - must be present
- "concept_id": the SNOMED CT concept ID as a string - this field is mandatory for every entry. If you are uncertain or a concept id is not found, provide 000000000 as concept ID. Never omit this field.

Example format:
<json>
[{"text": "airway", "start": 1078, "end": 1084, "concept_id": "89187006"}, {"text": "Penicillins", "start": 1168, "end": 1179, "concept_id": "764146007"}]
</json>

Be thorough and ensure every object in the array contains all four keys. Include all medical terms: anatomical sites, diagnoses, symptoms, procedures, medications, lab values, and clinical observations. "start" and "end" fields must be integers to denote character offsets.

\end{tcblisting}
\caption{Prompt template for the \textbf{End-to-End} setting.}
\label{fig:prompt2}
\end{figure*}

\begin{figure*}[h]
\centering
\begin{tcblisting}{
    colback=gray!10,
    colframe=black,
    title=Mention-based linking Prompt Template,
    width=0.95\textwidth, % Use almost full width
    listing only,         % Tells latex this is verbatim content
    listing options={
        basicstyle=\ttfamily\footnotesize, % Smaller monospaced font
        breaklines=true,       % AUTOMATICALLY WRAPS LONG LINES
        breakatwhitespace=false,
        columns=fullflexible,  % Better spacing
        keepspaces=true
    }
}

You are a medical terminology expert specializing in SNOMED CT (Systematized Nomenclature of Medicine - Clinical Terms).

Given a clinical note, your task is to:
1. Identify all medical entities (anatomical structures, clinical findings, procedures, substances, disorders, etc.)
2. For each entity, provide the exact mention text as it appears in the note, the SNOMED CT concept ID, and which occurrence of that text in the note it refers to.

<clinical_note>
{{ note_text }}
</clinical_note>

## Instructions
Extract every medical entity from the clinical note and link each to its SNOMED CT concept ID.

Reason about the correct mention and it's SNOMED CT concept inside <scratchpad> tags BEFORE you start writing JSON. Your <scratchpad> must be short and concise, in very few words. Once you begin the JSON array, do NOT interrupt it with any commentary, reasoning, or corrections - output the complete array in one unbroken block inside the <json> tags.

Respond with ONLY a JSON array, no other text. Each element must have:
- "text": the exact entity mention as it appears in the note (preserve original casing)
- "concept_id": the SNOMED CT concept ID as a string
- "occurrence": which occurrence of this exact text in the note (1 = first, 2 = second, etc.). Count occurrences case-insensitively.

For example, if "chest" appears 4 times in the note and the 3rd occurrence refers to SNOMED CT concept 51185008:
<json>
[{"text": "chest", "concept_id": "51185008", "occurrence": 3}]
</json>

Be thorough. Include all medical terms: anatomical sites, diagnoses, symptoms, procedures, medications, lab values, and clinical observations. If the same mention appears multiple times and each should be linked, include a separate entry for each occurrence.

\end{tcblisting}
\caption{Prompt template for the \textbf{Mention} setting.}
\label{fig:prompt3}
\end{figure*}

\begin{figure*}[h]
\centering
\begin{tcblisting}{
    colback=gray!10,
    colframe=black,
    title=Mention Extraction Prompt Template for RAG,
    width=0.95\textwidth, % Use almost full width
    listing only,         % Tells latex this is verbatim content
    listing options={
        basicstyle=\ttfamily\footnotesize, % Smaller monospaced font
        breaklines=true,       % AUTOMATICALLY WRAPS LONG LINES
        breakatwhitespace=false,
        columns=fullflexible,  % Better spacing
        keepspaces=true
    }
}

You are a medical terminology expert specializing in SNOMED CT (Systematized Nomenclature of Medicine - Clinical Terms).

Given a clinical note, your task is to identify all medical entities (anatomical structures, clinical findings, procedures, substances, disorders, etc.) and return each as a mention with its occurrence number.

<clinical_note>
{{ note_text }}
</clinical_note>

## Instructions
Extract every medical entity from the clinical note. Do NOT attempt to predict SNOMED CT concept IDs - only extract the mention text and occurrence number.

Reason about the correct mentions inside <scratchpad> tags BEFORE you start writing JSON. Your <scratchpad> must be short and concise, in very few words. Once you begin the JSON array, do NOT interrupt it with any commentary, reasoning, or corrections - output the complete array in one unbroken block inside the <json> tags.

Respond with ONLY a JSON array, no other text. Each element must have:
- "text": the exact entity mention as it appears in the note (preserve original casing)
- "occurrence": which occurrence of this exact text in the note (1 = first, 2 = second, etc.). Count occurrences case-insensitively.

For example, if "chest" appears 4 times in the note and the 3rd occurrence is a medical entity:
<json>
[{"text": "chest", "occurrence": 3}]
</json>

Be thorough. Include all medical terms: anatomical sites, diagnoses, symptoms, procedures, medications, lab values, and clinical observations. If the same mention appears multiple times and each should be linked, include a separate entry for each occurrence.

\end{tcblisting}
\caption{Prompt template for the \textbf{Mention extraction} (RAG Pt. 1).}
\label{fig:prompt4}
\end{figure*}

\begin{figure*}[h]
\centering
\begin{tcblisting}{
    colback=gray!10,
    colframe=black,
    title=Retrived-concept Linking Prompt Template for RAG,
    width=0.95\textwidth, % Use almost full width
    listing only,         % Tells latex this is verbatim content
    listing options={
        basicstyle=\ttfamily\footnotesize, % Smaller monospaced font
        breaklines=true,       % AUTOMATICALLY WRAPS LONG LINES
        breakatwhitespace=false,
        columns=fullflexible,  % Better spacing
        keepspaces=true
    }
}

You are a medical terminology expert specializing in SNOMED CT (Systematized Nomenclature of Medicine - Clinical Terms).

Given a clinical note and a list of medical entity mentions (each with candidate SNOMED CT concepts), your task is to select the best matching SNOMED CT concept ID for each mention from its candidate list.

<clinical_note>
{{ note_text }}
</clinical_note>

## Mentions and Candidates
{% for mention in mentions %}
### Mention {{ mention.id }}: "{{ mention.text }}"
Candidates:
{% for c in mention.candidates %}
- {{ c.snomed_id }}: {{ c.fsn }}
{% endfor %}
{% endfor %}

## Instructions
For each mention, select the single best-matching SNOMED CT concept from its candidate list. Consider the clinical context in the note to disambiguate.

Reason about each mention inside <scratchpad> tags BEFORE you start writing JSON. Your <scratchpad> must be short and concise. Once you begin the JSON array, do NOT interrupt it with any commentary - output the complete array in one unbroken block inside the <json> tags.

Respond with a JSON array with exactly {{ mentions|length }} elements, one per mention. Each element must have:
- "mention_id": the mention number (integer)
- "concept_id": the selected SNOMED CT concept ID as a string

If none of the candidates are appropriate, use "000000000" as the concept_id.

<json>
[{"mention_id": 0, "concept_id": "89187006"}, ...]
</json>

\end{tcblisting}
\caption{Prompt template for the \textbf{Concept selection} (RAG Pt. 2).}
\label{fig:prompt5}
\end{figure*}

\end{document}